\documentclass[10pt, a4paper]{article}
\usepackage[]{lrec-coling2024} % this is the new style

\usepackage[draft]{changes}
\definechangesauthor[color=red]{g}
\usepackage{float}
\usepackage{booktabs}
\usepackage{multirow}
\usepackage{graphicx}
\usepackage{amsmath}

\title{Prompting Explicit and Implicit Knowledge for Multi-hop Question Answering Based on Human Reading Process \thanks{This paper has been accepted at COLING 2024}}

\name{Guangming Huang\textsuperscript{1}, Yunfei Long\textsuperscript{1}, Cunjin Luo\textsuperscript{1}, Jiaxing Shen\textsuperscript{2}, Xia Sun\textsuperscript{3}} 

\address{\textsuperscript{1}School of Computer Science and Electronic Engineering, University of Essex, United Kingdom \\
\textsuperscript{2}Department of Computing and Decision Sciences, Lingnan University, Hong Kong, China \\
\textsuperscript{3}School of Information Science and Technology, Northwest University, China \\ 
\{gh22231, yl20051, cunjin.luo\}@essex.ac.uk, raindy@nwu.edu.cn, jiaxingshen@ln.edu.hk\\
}

\abstract{
Pre-trained language models (PLMs) leverage chains-of-thought (CoT) to simulate human reasoning and inference processes, achieving proficient performance in multi-hop QA. However, a gap persists between PLMs' reasoning abilities and those of humans when tackling complex problems. Psychological studies suggest a vital connection between explicit information in passages and human prior knowledge during reading. Nevertheless, current research has given insufficient attention to linking input passages and PLMs' pre-training-based knowledge from the perspective of human cognition studies. In this study, we introduce a \textbf{P}rompting \textbf{E}xplicit and \textbf{I}mplicit knowledge (PEI) framework, which uses prompts to connect explicit and implicit knowledge, aligning with human reading process for multi-hop QA. We consider the input passages as explicit knowledge, employing them to elicit implicit knowledge through unified prompt reasoning. Furthermore, our model incorporates type-specific reasoning via prompts, a form of implicit knowledge. Experimental results show that PEI performs comparably to the state-of-the-art on HotpotQA. Ablation studies confirm the efficacy of our model in bridging and integrating explicit and implicit knowledge.
 \\ \newline  }

\begin{document}

\maketitleabstract

\section{Introduction}

Multi-hop question answering (QA) represents a challenging task that demands intricate reasoning and inference across diverse sources to predict a coherent and precise answer \citep{yang2018hotpotqa,cao2023rpa}. Chain-of-thought (CoT) emulates the process of human reasoning and inference to generate a sequence of intermediate natural language reasoning steps that lead to the final results for complex reasoning problems. By employing prompt-based learning with CoT on pre-trained language models (PLMs), several studies have demonstrated proficient performance for multi-hop QA \citep{trivedi2022interleaving, deng2022prompt,zhang2023beam}.

However, a substantial disparity remains evident between the reasoning abilities of PLMs and human cognition when it comes to addressing complex problems. Current studies overlook the establishment of a direct link between the input passages and the knowledge assimilated by PLMs during the pre-training phase, considering the cognitive framework of humans \citep{wang2022iteratively,deng2022prompt,atif2023beamqa,cao2023rpa}.

In human reading comprehension studies, \citet{smith2012understanding} believes that information sources are often repeated during reading comprehension, leading to redundancy at linguistic levels, including letter-to-letter, word-to-word, sentence-to-sentence, and text-to-text. Consequently, readers can reduce their reliance on explicit information within the reading text by incorporating alternative sources of information, such as world knowledge \citep{hagoort2004integration}. According to \citet{clarke1977toward}, readers engage in the comprehension and question-answering process while reading, drawing upon both the explicit information conveyed in the text and their pre-existing language knowledge, background knowledge, and world knowledge derived from that explicit information. Certain studies have pointed out that a critical factor in reading ability is what the reader brings to the text, or what is usually referred to as prior knowledge \citep{yin1985role,baldwin1985effects,abdelaal2014relationship}. The experimental results demonstrate a significantly high correlation between the high prior knowledge and reading comprehension \citep{abdelaal2014relationship}.

\begin{figure*}[htb]
\centering 
\includegraphics[width=0.85\textwidth]{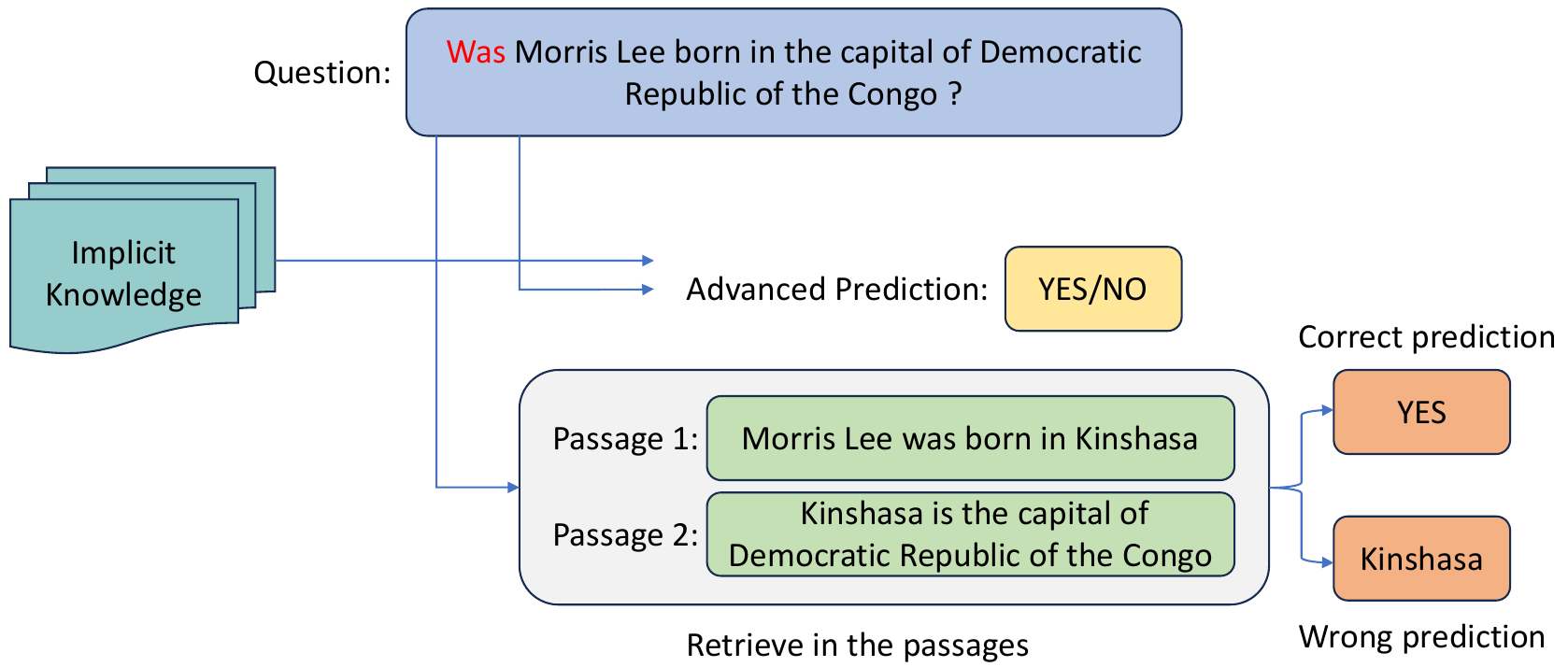}
\caption{An example of the significance of implicit knowledge in reading comprehension.} 
\label{Figure3} 
\end{figure*}

As an example shown in Figure \ref{Figure3}, given a question "\textit{Was Morris Lee born in the capital of Democratic Republic of the Congo ?}", human retrieves the information in the given passages to make a prediction. According to the auxiliary verb "\textit{was}" in the yes-no question, human can predict the answer as "\textit{yes}" or "\textit{no}" beforehand, drawing upon linguistic knowledge (as part of implicit knowledge), even in the absence of information regarding the capital of Congo.

% Figure 3

Consequently, an inseparable and inherent linkage prevails between the explicit information within the passages and the pre-existing prior knowledge. The prior knowledge diminishes the reliance on explicit information, thereby reducing the need for its extensive utilization. Moreover, the harmonious integration of prior knowledge and explicit information enhances the efficacy of the reading process, leading to improved comprehension and engagement. 

Incorporating insights from the aforementioned human cognition theories, we propose a novel framework called \textbf{P}rompting \textbf{E}xplicit and \textbf{I}mplicit knowledge (PEI) to address the challenges in multi-hop QA. Within this framework, readers are analogized to PLMs, their prior knowledge represents implicit knowledge acquired during the pre-training, and the explicit information in the passages serves as the input context conveying explicit knowledge. While acknowledging existing disparities between language models and human, and recognizing potential biases in considering readers as language models, \citet{jin2023evidence} believe that language models transcend being mere "stochastic parrots" \citep{bender2021dangers}, possessing the capability to acquire semantics during the pre-training.

To effectively utilize these knowledge sources, we employ prompts to capture explicit knowledge and invoke implicit knowledge. This facilitates the bridging of these two types of knowledge, thereby improving the performance of our proposed PEI model for multi-hop QA tasks. Moreover, this approach reduces the reliance on explicit knowledge present in the input passages by allowing selective filtering of irrelevant information or "redundancy" unrelated to the corresponding questions, as per \citet{smith2012understanding}'s theory. To further substantiate the contribution of implicit knowledge to our model's performance, we conduct an ablation study, the results of which affirm our hypothesis (see Section \ref{sec:ablation}).

\begin{figure*}[htb]
\centering 
\includegraphics[width=1\textwidth]{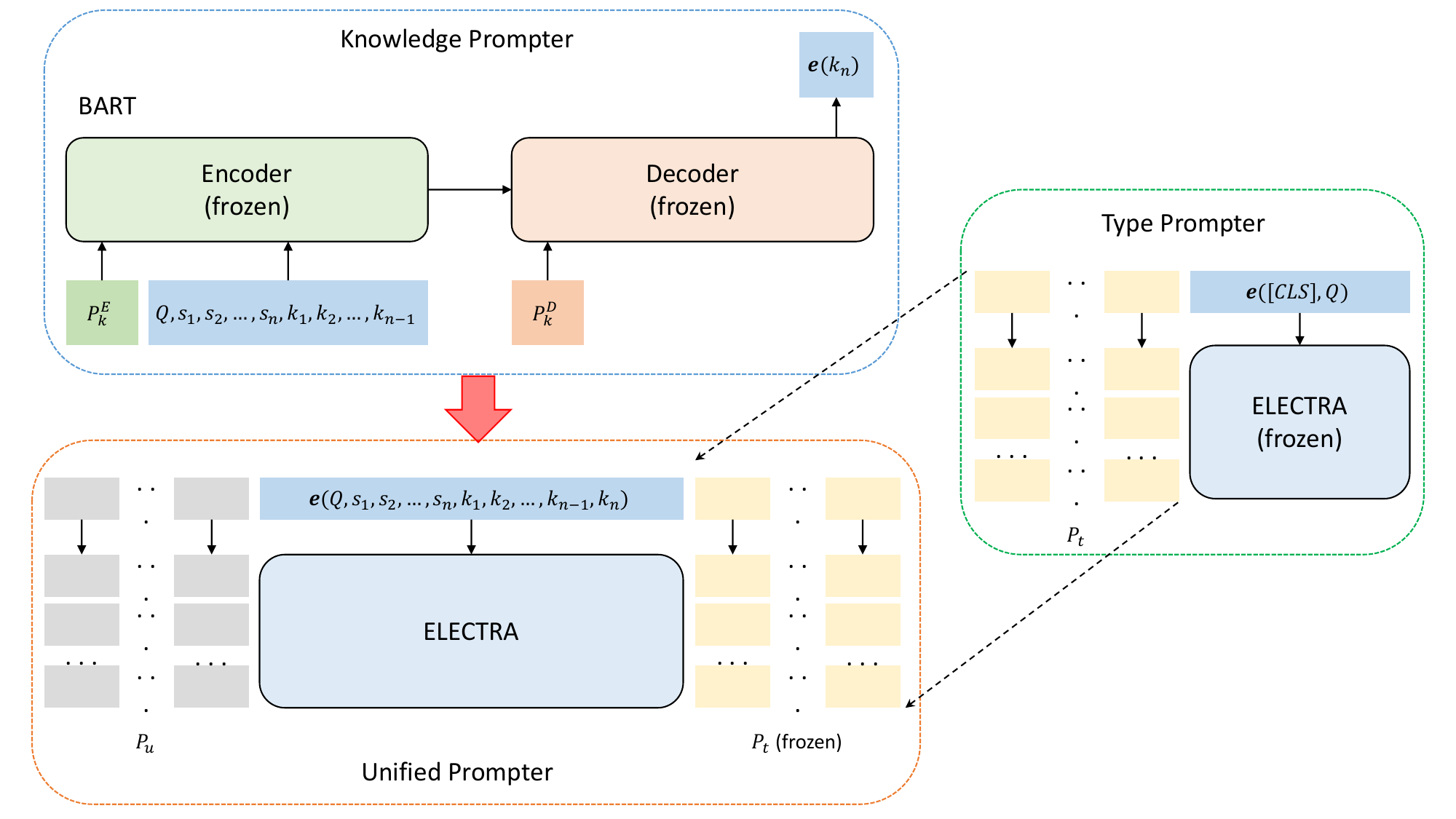}
\caption{The overview of our proposed PEI framework for multi-hop QA. The right green dashed block is the type prompter; the top blue dashed block refers to the knowledge prompter; and the bottom orange dashed block is the unified prompter.} 
\label{Figure1} 
\end{figure*}

Illustrated in Figure \ref{Figure1}, our proposed PEI framework consists of three components: 1) a type prompter for identifying and learning the weights of reasoning types for multi-hop questions; 2) an encoder-decoder PLM is employed to acquire implicit knowledge by leveraging explicit knowledge, mirroring the cognition process of human reading process.; 3) a unified prompt-based PLM integrating both explicit, implicit knowledge and question types for multi-hop QA.

Our contributions are summarized as follows:
\begin{itemize}
  \item The proposed PEI framework provides an effective approach for multi-hop QA based on human reading process, by modeling the input passages or context as explicit knowledge and PLMs mirroring with human prior knowledge as implicit knowledge.

  \item Our proposed PEI model achieves comparable performance with state-of-the-art on HotpotQA that is a benchmark dataset for multi-hop QA. Notably, our approach also maintains robust performance on corresponding single-hop sub-questions and other multi-hop datasets.

    \item Ablation studies confirm that implicit knowledge enhances the model's reasoning ability, supporting our hypothesis for the PEI model, grounded in the human reading process.
\end{itemize}

% Figure 1

\section{Related Work}
\subsection{Chain-of-Thought Prompting}
Prompt tuning has been recognized as an effective mechanism for conditioning PLMs to utilize relevant knowledge for specific downstream tasks \citep{lester2021power,liu2023pre}. CoT prompting, a prompt-based approach, has emerged as a method to recall implicit knowledge from large-scale language models (LLMs) for complex reasoning. It emulates the sequential and coherent reasoning process of human thinking by generating intermediate natural language reasoning steps that lead to the final outcome \citep{chu2023survey}. 
Manual-CoT \citep{wei2022chain} is designed to elicit the CoT reasoning ability through manual demonstrations. Subsequently, \citet{kojima2022large} demonstrated that LLMs can serve as competent zero-shot reasoners, as they generate rationales that already embody CoT reasoning, by incorporating the incantation "Let's think step by step" to facilitate a step-by-step thinking process before answering questions. An automatic CoT prompting, AutoCoT \citep{zhang2022automatic}, leverages diverse question sampling and reasoning chain generation to construct demonstrations, reducing the needs for human effort. Some studies explored CoT prompt learning for multihop QA \citep{wang2022iteratively, trivedi2022interleaving}. Inspired by these advancements, our work explores the utilization of CoT prompting to elicit implicit knowledge from PLMs. In contrast to CoT's generation of intermediate natural language steps, our model generates continuous embeddings as representations of implicit knowledge.

\subsection{Prompt-based Learning for Multi-hop QA}
Recent studies have made significant progress by employing prompts for multi-hop QA. For example, PromptRank \citep{khalifa2023few} constructs an instruction-based prompt that includes a candidate document path and then computes the relevance score between a given question and the path based on the conditional likelihood of the question given the path prompt according to a language model. IRCoT \citep{trivedi2022interleaving} interleaves CoT generation and knowledge retrieval steps to guide the retrieval using CoT prompting.
\citet{wang2022iteratively} introduced an iterative CoT prompting framework, which elicits knowledge from PLMs step by step utilizing a sequence-to-sequence BART-large model to recall a sequence of natural language statements for multi-hop QA. They employed a simple template to convert each triplet (subject entity, relation, object entity) in the evidence path into a natural language statement, which collectively formed the final statement. Inspired by their work, our approach also utilizes an encoder-decoder architecture PLM, i.e. BART-large, to recall implicit knowledge through iterative prompting. However, our proposed method differs distinctly from their study in three aspects: 1) our method does not necessitate the conversion of triple evidence paths into natural language statements; 2) we leverage input passages as explicit knowledge to elicit implicit knowledge from the PLM, which was not explored in their approach; 3) the recalled implicit knowledge in our model is represented as continuous embeddings rather than natural language statements or lexical knowledge \cite{huang2023lida}. Hence, our model does not rely on the natural language statements derived from the evidence paths.

\section{Methodology}
\subsection{Overview}
Shown in Figure \ref{Figure1}, our PEI framework comprises three components: 1) the type prompter identifies and acquires the weights associated with specific reasoning types for multi-hop questions; 2) the knowledge prompter elicits implicit knowledge through harnessing of explicit knowledge; 3) the unified prompt-based PLM integrates explicit and implicit knowledge, as well as question types, providing a comprehensive approach for multi-hop QA.

\textbf{Pre-training on Single-hop QA.} In order to understand the performance characteristics of current QA models at each step of the reasoning process, we built our works on a PLM foundation model, i.e. ELECTRA \footnote{The Pre-trained language models can also be replaced by more competent models. In line with previous works on prompt learning, we choose ELECTRA.} \citep{clark2020electra} on SQuAD \citep{rajpurkar2016squad}, which is a single-hop QA dataset. Subsequently, the pre-trained ELECTRA  model is employed as the backbone PLM for our type prompter module. By leveraging the trained ELECTRA model on single-hop, we aim to explore the interaction between the model's behaviour and the reasoning process across various hops.  

\subsection{Type Prompter}
The type prompter is crafted to facilitate the training process for the acquired weights of soft prompts, enabling them to effectively capture the distinctive features associated with various question types. As shown in Figure \ref{Figure1}, yellow blocks $P_t$ refer to trainable prompt embeddings; blue blocks are embeddings stored and frozen PLM.

Given question $Q$, we construct an input sequence consisting of soft prompts denoted as $\{P_t, [CLS], Q\}$, where $P_t$ represents the trainable soft prompts of the type prompter module. P-tuning v2 \citep{liu2021p} is employed to train soft prompts $P_t$ acquiring the weights of specific-type information of the given questions. As shown in Type Prompter of Figure \ref{Figure2}, we initially freeze the PLM and optimize the trainable soft prompt $P_t$. After training, we obtain the trained $P_t$ and establish a connection between it and the unified prompter, ensuring its fixed nature throughout subsequent operations.

We denote $d$ as the embedding dimension of the language model ELECTRA, $h$ as the number of layers within the PLM, and $l$ as the length of the prompt tokens. In this component, the total number of trainable parameters can be calculated as $\Theta(d\cdot h\cdot l)$. Compared with the fine-tuning paradigm, the type prompter module based on p-tuning can reduce training parameters while ensuring that it can capture the type-specific information. Secondly, it facilitates to transfer the trained weights of $P_t$ with the type-specific information to unified prompter module. Moreover, through the application of the p-tuning v2, we can effectively capture and learn a broader range of features than prompt tuning \citep{lester2021power}.

\subsection{Knowledge Prompter}
By capitalizing on the textual information input, the knowledge prompter seeks to activate and engage the innate prior knowledge possessed by individuals, thereby facilitating the assimilation of these two information sources for comprehensive reading comprehension. 
As illustrated in Figure \ref{Figure1}, the knowledge prompter, an iterative encoder-decoder PLM, retrieves implicit knowledge through prefix tuning \citep{li2021prefix}. The trainable prompt embeddings, denoted as $P_k^E$ and $P_k^D$, are integrated into each layer of the encoder and decoder of the PLM, respectively. This incorporation of prompt embeddings facilitates the efficient retrieval and utilization of explicit knowledge throughout the iterative encoding and decoding phases.

Given a multi-hop query $Q$ and a sequence of supporting sentences $S_n = [s_1, s_2,..., s_i,..., s_n]$, our objective is to retrieve a sequence of knowledge $K_n = [k_1, k_2,.., k_i,..., k_n]$ that provides sufficient information for determining the response to both $Q$ and $S_n$, where $n$ represents the number of supporting sentences. Our focus lies in the development of prompting methods, where we aim to construct prompts $P_k^E$ and $P_k^D$ to guide the encoder-decoder language model in recalling the desired knowledge $K_n$. Notably, we maintain fixed parameters for the encoder-decoder PLM, allowing us to direct its retrieval process through the strategic construction of prompts.

Motivated by the sequential nature observed in multi-step reasoning tasks \citep{wang2022iteratively}, we adopt an iterative approach as below: 

\begin{equation} \label{eq:1}
\begin{aligned}
P(k_j | Q, S_j, K_{j-1}) = \prod_{j=1}^{n}P(k_j | Q, s_1,...,& s_{j}, \\
k_1,...,k_{j-1}) 
\end{aligned}
\end{equation}

\begin{equation} \label{eq:2}
\begin{aligned}
decoder(k_j) = encoder(Q, S_j,K_{j-1})
\end{aligned}
\end{equation}

where at each step $j$, PLM recalls the next piece of knowledge $k_j$ conditioned on the query $Q$ and supporting sentences $s_1,..., s_{j}$ and gathered knowledge $k_1,...,k_{j-1}$.

More specially, when $j = 1$, it is written as following based on Equation (\ref{eq:1}) and (\ref{eq:2}):

\begin{equation} \label{eq:3}
\begin{aligned}
decoder(k_1) = encoder(Q, s_1)
\end{aligned}
\end{equation}

\subsection{Unified Prompter}
As described in Figure \ref{Figure1}, we concatenate the unified prompter module and $P_t$, which is with the weights of type-specific reasoning. Moreover, the implicit knowledge $K_n$ from the knowledge prompter module is the additional inputs. Intuitively, this amalgamation of two information is capable to enhances the model's reasoning ability. 

In the unified prompter, the trainable prompt embeddings are denoted as $P_u$, while we freeze the trained promopt embeddings $P_t$. To preserve the learned weights of $P_t$ from the type prompter, we adopt the identical architecture of the language model (ELECTRA), allowing for seamless concatenation of $P_t$ with the unified prompter. Subsequently, we perform joint fine-tuning and prompt-tuning of both the language model and the prompt $P_u$ to optimize the overall performance of the model in the context of multi-hop question-answering (QA) prediction.

\section{Experiments}

\subsection{Dataset and Metrics}

\textbf{HotpotQA} \citep{yang2018hotpotqa} contains a collection of 113k question-answer pairs drawn from Wikipedia. Additionally, HotpotQA offers sentence-level supporting facts that are essential for the process of reasoning, which allows QA systems to infer with robust supervision and explain the predictions.

\textbf{2WikiMultiHopQA} \citep{ho2020constructing} has over 192k samples, including 167k training, 12.5k evaluation, and 12.5 test samples. The format of the dataset primarily follows HotpotQA \citep{yang2018hotpotqa}, but it provides additional enhancements, including a broader range of reasoning types for questions and comprehensive annotations of evidence paths associated with each question.

\textbf{MuSiQue} \citep{trivedi2022musique} contains 25k 2-4 hop questions samples. This dataset involves the systematic selection of composable pairs of single-hop questions demonstrating logical connections to generate a set of multi-hop questions.

\textbf{Sub-question QA dataset} \citep{tang2021multi} was created to facilitate the analysis of the reasoning capabilities of multi-hop QA models at each step of the reasoning process. To evaluate the performance of these models, the authors curated a specialized dataset consisting of single-hop sub-questions. This dataset comprises 1000 samples manually verified from the development set of HotpotQA, ensuring high-quality evaluation resource for  the study.

In order to ensure consistency and comparability across the datasets used in our experimental evaluations, we categorize the question types of the three datasets under investigation into the broader categories of comparison and bridge. This categorization of question types facilitates a standardized approach to handling diverse question structures across the datasets subjected to our evaluation.

\textbf{Metrics.} We employ Exact Match ($EM$) and Partial Match ($F1$) to evaluate the efficacy and performance of our proposed framework concerning both answer and supporting facts prediction. Furthermore, the joint $EM$ and $F1$  are used to assess the overall performance.

\subsection{Implementation Details}
Inspired by the studies of \citet{wang2022iteratively} and \citet{deng2022prompt}, we adopt BART-large \citep{lewis2020bart} as the language model in the knowledge prompter module. ELECTRA-large \citep{clark2020electra} serves as the foundation language model for both the type prompter and unified prompter modules. Our implementation is built upon the Huggingface platform \citep{wolf2020transformers}. For model optimization, we employ the AdamW optimizer \citep{loshchilov2018decoupled} along with a linear learning rate scheduler with a warmup ratio of $0.05$.

In terms of hyperparameters, we conduct a search for the optimal batch size. For the HotpotQA, 2WikiMultiHopQA, and MuSiQue datasets, we explored batch sizes of $\{4, 8, 12, 16, 32\}$ respectively. Additionally, we performed a tuning process for the learning rate, considering values from $\{2e-5, 4e-5, 8e-5, 2e-4, 4e-4, 8e-4, 2e-3, 4e-3, 8e-3, 2e-2, 4e-2, 8e-2\}$. Moreover, we conducted tuning experiments for the length of the encoder/decoder prompts $P_k$ and the type prompts $P_t$, exploring values from $\{15, 30, 45, 60, 75, 90, 100\}$.

\subsection{Main Results and Analysis}

% Table 1
\begin{table*}[htb]
\centering
\begin{tabular}{@{}lcccccc@{}}
\toprule
\multicolumn{1}{c}{\multirow{2}{*}{Models}} & \multicolumn{2}{c}{Ans} & \multicolumn{2}{c}{Sup}         & \multicolumn{2}{c}{Joint} \\ \cmidrule(l){2-7} 
\multicolumn{1}{c}{}                     & EM             & F1             & EM    & F1    & EM    & F1             \\ \midrule
Baseline Model \citep{yang2018hotpotqa}  & 45.60          & 59.02          & 20.32 & 64.49 & 10.83 & 40.16          \\
DecompRC \citep{min2019multi}            & 55.20          & 69.63          & -     & -     & -     & -              \\
OUNS \citep{perez2020unsupervised}       & 66.33          & 79.34          & -     & -     & -     & -              \\
QFE \citep{nishida2019answering}         & 53.86          & 68.06          & 57.75 & 84.49 & 34.63 & 59.61          \\
DFGN \citep{qiu2019dynamically}          & 56.31          & 69.69          & 51.50 & 81.62 & 33.62 & 59.82          \\
SAE-large \citep{tu2020select}           & 66.92          & 66.92          & 61.53 & 86.86 & 45.36 & 71.45          \\
C2F Reader \citep{shao2020graph}         & 67.98          & 81.24          & 60.81 & 87.63 & 44.67 & 72.73          \\
Longformer \citep{beltagy2020longformer} & 68.00          & 81.25          & 63.09 & 88.34 & 45.91 & 73.16          \\
HGN \citep{fang2020hierarchical}   & 69.22          & 82.19          & 62.76 & 88.47 & 47.11 & 74.21          \\
AMGN \citep{li2021asynchronous}          & 70.53          & 83.37          & 63.57 & 88.83 & 47.77 & 75.24          \\
S2G \citep{wu2021graph}                  & 70.72          & 83.53          & 64.30 & 88.72 & 48.60 & 75.45          \\
iCAP $^{\dag}$ \citep{wang2022iteratively}         & 68.61          & 81.82          & 62.80 & 88.51 & 47.02 & 74.11          \\
PCL \citep{deng2022prompt}               & 71.76          & 84.39          & 64.61 & 89.20 & 49.27 & 76.56          \\
Beam Retrieval \citep{zhang2023beam}        & \underline{72.69}      & \underline{85.04}      & \textbf{66.25} & \textbf{90.09} & \textbf{50.53}   & \underline{77.54}  \\ \midrule
PEI (Ours)     & \textbf{72.89} & \textbf{85.32} & \underline{65.03} & \underline{89.81} & \underline{49.91} & \textbf{77.84} \\ \bottomrule
\end{tabular}
\caption{Results on the blind test set of HotpotQA in the distractor setting. “-” denotes the case where no results are available. $^{\dag}$ denotes that we implement the codes. Other results are derived from \citep{deng2022prompt} and \citep{zhang2023beam}. “Ans” represents the metrics for answer; “Sup” denotes the metrics for supporting facts; “Joint” is the joint metrics that combine the evaluation of answer spans and supporting facts.}
\label{tab:1}
\end{table*}

Initially, we evaluate our proposed PEI model on the test set of HotpotQA in the distractor setting comparing with peer-reviewed baselines, including the baseline model of HotpotQA \citep{yang2018hotpotqa}, SOTA model on the leaderboard \cite{zhang2023beam}, iCAP \cite{wang2022iteratively} and PCL \cite{deng2022prompt} which we inspired by, and other baselines.
Illustrated in Table \ref{tab:1}, our proposed PEI model demonstrates superior performance across all evaluation metrics compared to all other baselines (except Beam Retrieval), and achieves comparable performance with Beam Retrieval \cite{zhang2023beam} on the HotpotQA dataset, highlighting the significant progress made by PEI for multi-hop QA.

More specifically, our PEI achieves an improvement of $0.20$ in answer EM, $0.28$ in answer F1 score, and $0.30$ in join F1 score when compared to Beam Retrieval. Conversely, Beam Retrieval shows an improvement of $1.22$ in supporting EM, $0.28$ in supporting F1 score, and $0.62$ in join EM when compared to PEI.

The difference in performance between PEI and Beam Retrieval in answer prediction versus supporting prediction can be attributed to their respective approaches. Beam Retrieval maintains multiple partial hypotheses of relevant passages at each step, expanding the search space (albeit at the expense of an exponentially complex retrieval process) and reducing the risk of missing relevant passages. Consequently, it excels in supporting prediction. In contrast, our PEI model leverages insights from the human reading process, incorporating implicit knowledge and type-specific information. This approach contributes to the model's improvement in answer prediction. However, it may not exhibit the same level of performance in supporting prediction as Beam Retrieval due to differences in retrieval strategies.

Moreover, our PEI framework demonstrates a significant improvement of $0.64/1.28$ in the Joint EM/F1 score compared to the PCL model, which PEI and PCL both use the same backbone PLM (i.e., ELECTRA). Although PCL also identified the reasoning type of multi-hop question as a soft prompt via a transformer-based question classifier, our proposed model not only consider the type-specific knowledge but also incorporate the implicit knowledge through iteratively eliciting it from an encoder-decoder PLM. Additionally, our proposed PEI outperforms iCAP with $2.89/3.73$ in the joint EM/F1 score, despite both PEI and iCAP utilizing the same encoder-decoder skeleton PLM (i.e.,BART). Compared to the graph-based model AMGN, our PEI framework exhibits substantial gains, with an improvement of $2.14/2.6$ in the Joint EM/F1 score. This indicates that our framework achieves better performance employing the same backbone PLM.

\subsection{Evaluation of Robustness}
Since our proposed model employs identical backbone models (i.e., ELECTRA and BART) and similar prompt learning framework as PCL \citep{deng2022prompt} and iCAP \citep{wang2022iteratively}, we further evaluate the robustness of our model, particularly in comparison to PCL and iCAP. Additionally, we extend our evaluation to include a graph-based model, HGN \citep{fang2020hierarchical}, to ensure a thorough assessment of its robustness.

\textbf{Evaluation on Other Multi-hop Datasets.} To assess the generalization, we evaluate PEI model on 2WikiMultihopQA and MuSiQue. In Table \ref{tab:2}, the results show that our PEI model surpasses all comparison baselines in terms of both EM and F1 metrics. Notably, despite both PEI and iCAP employ the same encoder-decoder BART model, PEI achieves a substantial improvement of $4.52/26.66$ in the answer EM/F1 score compared to iCAP on the 2WikiMultihopQA dataset. Furthermore, our model exhibits superior performance compared to PCL, yielding improvements of $1.29/1.14$ and $0.69/0.51$ in the answer EM/F1 score on 2WikiMultihopQA and MuSiQue, respectively.

% Table 2 
\begin{table}[htb]
\centering
% \resizebox{0.48\textwidth}{!}{%
\begin{tabular}{@{}lcccc@{}}
\toprule
\multirow{2}{*}{Models} & \multicolumn{2}{c}{2WikiMultihopQA}       & \multicolumn{2}{c}{MuSiQue}     \\ \cmidrule(l){2-5} 
                        & EM             & F1             & EM             & F1             \\ \midrule
iCAP                    & 42.80          & 47.90          & -              & -              \\
HGN                     & 38.74          & 68.69          & 39.42          & 65.12          \\
PCL                     & 46.03          & 73.42          & 41.28          & 67.34          \\ \midrule
PEI (Ours)              & \textbf{47.32} & \textbf{74.56} & \textbf{41.97} & \textbf{67.85} \\ \bottomrule
\end{tabular}%
% }
\caption{Results of our proposed PEI compared to PCL, HGN \citep{deng2022prompt} and iCAP \citep{wang2022iteratively} on 2WikiMultihopQA and MuSiQue multi-hop QA test set. “-” denotes the case where no results are available.}
\label{tab:2}
\end{table}

\begin{figure}[htb]
\centering 
\includegraphics[width=1\columnwidth]{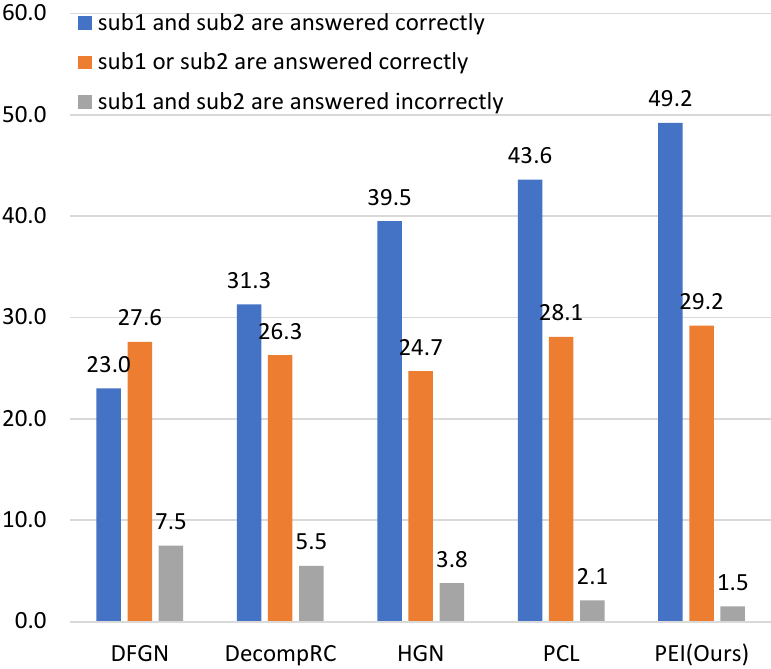} 
\caption{The success rate (\%) of five multi-hop QA models. $sub1$ denotes the first sub-question and $sub2$ is the second sub-question of corresponding question $q$. The results of DFGN, DecompRC, HGN and PCL are  from \citep{deng2022prompt}.} 
\label{Figure2} 
\end{figure}

\textbf{Evaluation on Sub-question Dataset.} To evaluate the PEI model's efficacy in the multi-hop reasoning process, specifically in composing answers from solved sub-questions, we conducted an evaluation on sub-question QA dataset \citep{tang2021multi}. The parent questions are denoted as $q$ along with their corresponding sub-questions $q_{sub1}$ and $q_{sub2}$. Table \ref{tab:3} shows that our PEI model achieves a $97.62\%$ success rate in correctly answering the parent multi-hop question $q$ when both sub-questions $q_{sub1}$ and $q_{sub2}$ are answered correctly \footnote{The calculation process: $49.2/(49.2+1.2)=97.62\%$}. This highlights the proficiency of PEI in retaining acquired knowledge through the integration of explicit and implicit knowledge, surpassing other multi-hop QA models. However, it is noteworthy that PEI also demonstrates a significant likelihood (36.55\%) of correctly answering the parent multi-hop question even when only one of the sub-questions is answered accurately \footnote{The calculation process: $(7.1+22.1)/(49.2+7.1+22.1+1.5)=36.55\%$}. To further illustrate the sub-question dependent success rate of various multi-hop QA models, Figure \ref{Figure2} shows that these models exhibit a high likelihood of correctly answering parent multi-hop questions even when only one sub-question is answered correctly. Therefore, in multi-hop QA, it is a prevalent and unsolved problem that models unexpectedly leverage unreliable reasoning shortcuts for answer predictions, which we do not expect that to happen \cite{deng2022prompt}.

% Table 3 
\begin{table*}[htb]
\centering
\begin{tabular}{@{}cccccccc@{}}
\toprule
$q$ & $q_{sub1}$ & $q_{sub2}$ & DFGN & DecompRC & HGN  & PCL  & PEI(Ours)     \\ \midrule
c & c       & c       & 23.0 & 31.3     & 39.5 & 43.6 & 49.2 \\
c & c       & w       & 9.7  & 7.2      & 5.1  & 6.8  & 7.1  \\
c & w       & c       & 17.9 & 19.1     & 19.6 & 21.3 & 22.1 \\
c & w       & w       & 7.5  & 5.5      & 3.8  & 2.1  & 1.5  \\ \midrule
w & c       & c       & 4.9  & 3.0      & 2.8  & 1.7  & 1.2  \\
w & c       & w       & 17.0 & 18.6     & 16.7 & 16.3 & 13.4 \\
w & w       & c       & 3.5  & 3.4      & 2.6  & 1.1  & 1.0  \\
w & w       & w       & 16.5 & 11.9     & 9.9  & 7.1  & 4.5  \\ \bottomrule
\end{tabular}
\caption{Results on sub-question dataset. $c/w$ denotes that the question is answered correctly/wrongly. $sub1$ denotes the first sub-question and $sub2$ is the second sub-question of corresponding question $q$. The results of DFGN, DecompRC, HGN and PCL are derived from \citep{deng2022prompt}.}
\label{tab:3}
\end{table*}

% Figure 2

\subsection{Ablation Studies} \label{sec:ablation}
To evaluate the contributions of distinct components within our PEI model, we conducted a series of ablation studies using the development set of HotpotQA as the experimental platform.

% Table 4
\begin{table*}[htb]
\centering
\begin{tabular}{@{}lccc@{}}
\toprule
\multicolumn{1}{c}{Model} & Ans F1         & Sup F1         & Joint F1       \\ \midrule
ELECTRA                   & 78.12          & 88.20          & 73.50          \\
- Type Prompter           & 81.14 $\uparrow_{3.02}$         & 89.37 $\uparrow_{1.17}$          & 75.68 $\uparrow_{2.18}$        \\
- Pre-trained             & 78.82 $\uparrow_{0.70}$          & 88.82 $\uparrow_{0.62}$         & 74.54 $\uparrow_{1.04}$         \\
- Implicit knowledge      & 81.22 $\uparrow_{3.10}$         & 90.25  $\uparrow_{2.05}$         & 75.80 $\uparrow_{2.30}$         \\ \midrule
PEI                       & \textbf{85.68} $\uparrow_{7.56}$ & \textbf{92.11} $\uparrow_{3.91}$ & \textbf{79.02} $\uparrow_{5.52}$ \\ \bottomrule
\end{tabular}%
\caption{Ablation Study of PEI on the development set of HotpotQA. Ans F1 stands for answer F1; Sup F1 is supporting F1.}
\label{tab:4}
\end{table*}

\textbf{Effect of Implicit Knowledge.}
To verify the hypothesis that implicit knowledge enhances reasoning for multi-hop QA, we conduct an ablation study between the performance of the ELECTRA model with and without implicit knowledge. In Table \ref{tab:4}, the experimental results show that the ELECTRA model with implicit knowledge improves $3.10/2.05/2.30$ in Ans F1/Sup F1/Joint F1 compared to the model without the implicit knowledge. These findings support the notion that implicit knowledge enhances the reasoning ability of the model thereby proving the hypothesis underlying our proposed PEI model, inspired by human reading comprehension.

\textbf{Effect of Type Prompts.}
To verify the effects of the type prompts and perform type-specific reasoning on multihop QA, we conduct a comparative analysis between the performance of the ELECTRA language model with and without the type prompter. As illustrated in Table \ref{tab:4}, the language model combined with the type prompter achieves a substantial improvement of $3.02/1.17/2.18$ in Ans/Sup/Joint F1 compared to the model without the type prompter component. These findings demonstrate that integrating question type information through the type prompter module effectively enhances the overall performance of the model and enables type-specific reasoning for multi-hop QA. Furthermore, these results validate the alignment of our model design with the cognitive processes observed in human reasoning. Because type information could be consider as one form of implicit knowledge. 

\textbf{Effect of Pre-training on Single-hop.}
Initially, we trained an ELECTRA-based QA model on the single-hop QA dataset SQuAD \citep{rajpurkar2016squad}, and subsequently retrained it on the HotpotQA dataset. Although conservation learning \citep{deng2022prompt} is not employed in our model, we evaluate the performance of our model with and without pre-training in order to verify the effect of the pre-training on single-hop. As shown in Table \ref{tab:4}, the language model combined with the pre-training improves $0.70/0.62/1.04$ in Ans F1/Sup F1/Joint F1 compared to the model without the pre-training. This indicates that pre-training in the single-hop QA task enable the model to acquire valuable information, enhancing its overall performance. However, it is noteworthy that this improvement is relatively limited without conservation learning.

\textbf{Effect of Foundation PLMs.}
To assess the effects of foundation PLMs, we compare PEI with PCL and HGN based on same experimental context including data and foundation models. Illustrated in Table \ref{tab:5}, PEI consistently outperforms PCL and HGN across all metrics. This indicates the effectiveness and robustness of PEI across PLMs. Moreover, PEI with ALBERT achieves an improvement of $0.62/0.23$ in Ans/Sup F1 compared to PEI with RoBERTa. ALBERT outperforms RoBERTa on GLUE benchmark using a single-model setup \footnote{https://github.com/google-research/albert}. These results confirm that adopting a more competent foundation PLM can improve the performance of our model.

% TAbel 5 
\begin{table}[htb]
\centering
\resizebox{\columnwidth}{!}{%
\begin{tabular}{@{}lccc@{}}
\toprule
\multicolumn{1}{c}{Model} & Ans F1         & Sup F1         & Joint F1       \\ \midrule
HGN (RoBERTa)             & 82.22          & 88.58          & 74.37          \\
HGN (ELECTRA)             & 82.24          & 88.63          & 74.51          \\
HGN (ALBERT)              & 83.46          & 89.20          & 75.79          \\ \midrule
PCL (RoBERTa)             & 84.33          & 90.75          & 77.12          \\
PCL (ELECTRA)             & 84.42          & 91.15          & 77.76          \\
PCL (ALBERT)              & 85.47          & 91.28          & 78.76          \\ \midrule
PEI (RoBERTa)             & 85.61          & 92.02          & 78.95          \\
PEI (ELECTRA)             & 85.68          & 92.11          & 79.02          \\
PEI (ALBERT)              & \textbf{86.23} & \textbf{92.25} & \textbf{79.11} \\ \bottomrule
\end{tabular}%
}
\caption{Results with different PLMs on the development set of HotpotQA. RoBERTa, ELECTRA and ALBERT represent RoBERTa-large (\#params: 355M), ELECTRA-large (\#params: 335M) and ALBERT-xxlargev2 (\#params: 223M)  as the PLM respectively. The results of HGN and PCL are derived from \citep{deng2022prompt}.}
\label{tab:5}
\end{table}

\section{Conclusions and Future Work}
In this paper, we introduce a novel framework that mimics human cognitive reading processes, employing prompts to bridge explicit and implicit knowledge. Our framework leverages prompts to elicit implicit knowledge from PLMs within the input context. Additionally, we integrate question type information to enhance model performance. Experimental results show that PEI performs comparably to the state-of-the-art on HotpotQA. Furthermore ablation studies confirm the effectiveness and robustness of our model in emulating human reading processes. In the future, we aim to extend and apply human reading cognition theories to diverse reasoning tasks, with the hope of enabling stronger, complex reasoning capabilities.

\section{Limitations}

\textbf{Human reading and cognition theory for reasoning.}
While our experimental results demonstrate that our model outperforms all but one baseline model for multi-hop QA, we have solely validated the efficacy of these cognition theories within this specific domain. Future research opportunities include extending these principles to diverse reasoning tasks, such as mathematical reasoning. Additionally, exploring alternative theories of human cognition and their potential applications in reasoning tasks would be valuable.

\textbf{Interpretability of implicit knowledge.} 
Although leveraging implicit knowledge benefits the model's reasoning process, the soft prompting we elicit are challenging to explain. Currently, it is a lack of insights into the specific knowledge acquired and how it contributes to the model's reasoning processes and decision-making capabilities.

\textbf{Experiments with larger-scale models.} While we currently conduct our experiments with BART and ELECTRA, the availability of large-scale language models like GPT-3 presents an opportunity to enhance the model's capabilities by incorporating more powerful models with richer prior knowledge. Nevertheless, it remains imperative to strike a balance between computational cost and model performance.

\section{Acknowledgements}

This work is supported by the Alan Turning Institute/DSO grant: \textit{Improving multimodality misinformation detection with affective analysis}. Yunfei Long, Guangming Huang and Cunjin Luo acknowledge the financial support of the School of Computer science and Electrical Engineering, University of Essex.

\section{Bibliographical References}\label{sec:reference}
\bibliographystyle{lrec-coling2024-natbib}
\bibliography{lrec-coling2024-example}

\end{document}